\documentclass[conference]{IEEEtran}
\IEEEoverridecommandlockouts

\usepackage{cite}
\usepackage{amsmath,amssymb,amsfonts}
\usepackage{comment}
\usepackage{algorithm}
\usepackage{algpseudocode}

\usepackage{float}
\usepackage{multirow}
\usepackage{makecell}
\usepackage{caption}
\usepackage{subcaption}
\usepackage{graphicx}
\usepackage{textcomp}
\usepackage{xcolor}
\usepackage{setspace}
\newcommand*{\affaddr}[1]{#1} 
\newcommand*{\affmark}[1][*]{\textsuperscript{#1}}
\newcommand*{\email}[1]{\texttt{#1}}

\usepackage{mathtools}

\DeclarePairedDelimiter\floor{\lfloor}{\rfloor}

\makeatletter
\def\blfootnote{\xdef\@thefnmark{}\@footnotetext}
\makeatother

\def\BibTeX{{\rm B\kern-.05em{\sc i\kern-.025em b}\kern-.08em
    T\kern-.1667em\lower.7ex\hbox{E}\kern-.125emX}}
    
\begin{document}

\title{SGQuant: Squeezing the Last Bit on Graph Neural Networks with Specialized Quantization}

\author{
 Boyuan Feng*, Yuke Wang*, Xu Li, Shu Yang, Xueqiao Peng, and Yufei Ding\\  
\affaddr{University of California, Santa Barbara, USA} \\
\email{\{boyuan,yuke\_wang,shuyang1995,yufeiding\}@cs.ucsb.edu} \\ 
\email{\{lixu9906,ameliapxq0131\}@gmail.com}\\
}

\maketitle

\blfootnote{*The first two authors contribute equally.}
 \pdfoutput=1

{\setstretch{0.915}
\begin{abstract}
With the increasing popularity of graph-based learning, Graph Neural Networks (GNNs) win lots of attention from research and industry field because of their high accuracy. However, existing GNNs suffer from high memory footprints (\textit{e.g.}, node embedding features).
This high memory footprint hurdles the potential applications towards memory-constrained devices, such as the widely-deployed IoT devices.
To this end, we propose a specialized GNN quantization scheme, SGQuant, to systematically reduce the GNN memory consumption.
Specifically, we first propose a GNN-tailored quantization algorithm design and a GNN quantization fine-tuning scheme to reduce memory consumption while maintaining accuracy.
Then, we investigate the multi-granularity quantization strategy that operates at different levels (components, graph topology, and layers) of GNN computation. 
Moreover, we offer an automatic bit-selecting (ABS) to pinpoint the most appropriate quantization bits for the above multi-granularity quantizations.
Intensive experiments show that SGQuant can effectively reduce the memory footprint from 4.25$\times$ to 31.9$\times$ compared with the original full-precision GNNs while limiting the accuracy drop to 0.4\% on average.
\end{abstract}

\section{Introduction}
Recently, Graph Neural Networks (GNNs) emerge as a new tool to manage various graph-based deep learning tasks (\textit{e.g.,} node classification~\cite{kaspar2010graph, gibert2012graph, duran2017learning} and link prediction~\cite{chen2005link, kunegis2009learning, tylenda2009towards}). In the comparison with standard methods for graph analytics, such as random walk~\cite{grover2016node2vec, deepWalk} and graph laplacians~\cite{luo2011cauchy, luo2009non, cheng2018deep}, GNNs highlight themselves with significantly higher accuracy~\cite{GCNConv, GINConv, GATConv} and better generality~\cite{SageConv}. 
In addition, the well-learned GNNs~\cite{GCNConv, GINConv, SageConv, GATConv} can be easily applied towards different types of graph structures or dynamic graphs without much re-computing overhead.

However, the GNNs featured with high memory footprint prevent them from being effectively applied towards the vast majority of resource-constrained settings, such as embedded systems and IoT devices, which are essential for many domains. There are two major reasons behind such an awkward situation. First, the input of GNNs consists of two types of inputs, \textit{graph structures} (edge list) and \textit{node features} (embeddings), which would easily lead to a dramatic increase in their storage sizes when the graph becomes large. This will stress the very limited memory budgets of those small devices. Second, the larger size of graphs demands more data operations (\textit{e.g.}, addition and multiplication) and data movements (\textit{e.g.}, memory transactions), which will consume lots of energy and drain the limited power budget on those tiny devices.
To tackle these challenges, data quantization can emerge as an ``one-stone-two-bird'' solution for resource-constrained devices that can 1) effectively reduce the memory size of both the graph structure and node embeddings, leading to less memory usage; 2) effectively minimize the size of manipulated data, leading to less power consumption. 

Nevertheless, an efficient approach for GNN quantization is still missing. Existing approaches may 1) choose a simple yet aggressive uniform quantization to all data to minimize memory and power cost, which leads to high accuracy loss; 2) choose a very conservative quantization to maintain accuracy, which leads to sub-optimal memory and energy-saving performance. While numerous works have been explored for quantization on CNNs~\cite{han2015deep, banner2019post, zhu2016trained, lin2016fixed}, directly applying these existing techniques without considering GNN-specific properties, would easily result in unsatisfactory quantization performance. To address these problems, we believe that three critical questions are noteworthy: 1) what types of data (weight or features) should be quantized? 2) what is the efficient quantization scheme suitable for GNNs? 
3) How to determine the quantization bits?
\begin{figure}[h] \small
    \centering
    \vspace{-10pt}
    \includegraphics[width=0.9\columnwidth]{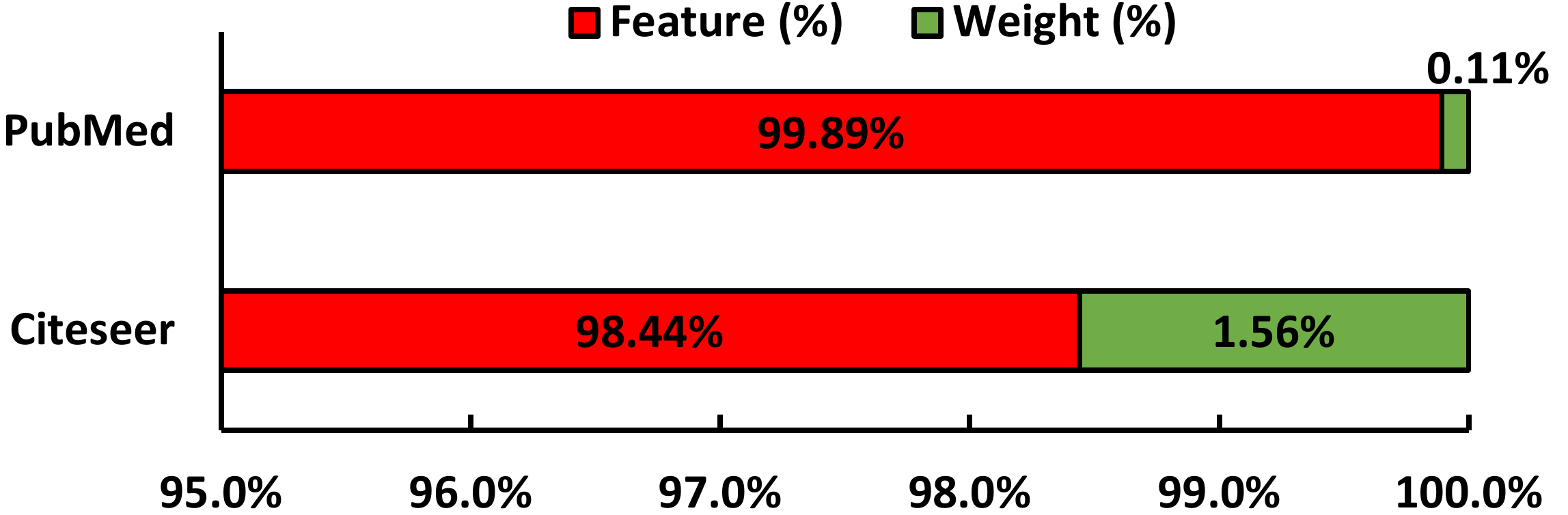}
    \caption{GAT Feature/Weight Memory Size Ratio.}
    \vspace{-5pt}
    \label{fig: Feature/Weight Memroy Size Ratio.}
\end{figure} 

To answer these questions, we make the following observations: a) quantization on node embedding features is more effective. As shown in Figure~\ref{fig: Feature/Weight Memroy Size Ratio.}, the features take up to $99.89\%$ of the overall memory size, which demonstrates their significant memory impact; b) GNNs computing paradigms are different across different layers, different graphs nodes, different components. And these differences could be leveraged as the major "guideline" for enforcing more efficient character-driven quantization.
\begin{figure}[t] \small
    \centering
    \vspace{-10pt}
    \includegraphics[width=0.9\columnwidth, height=120pt]{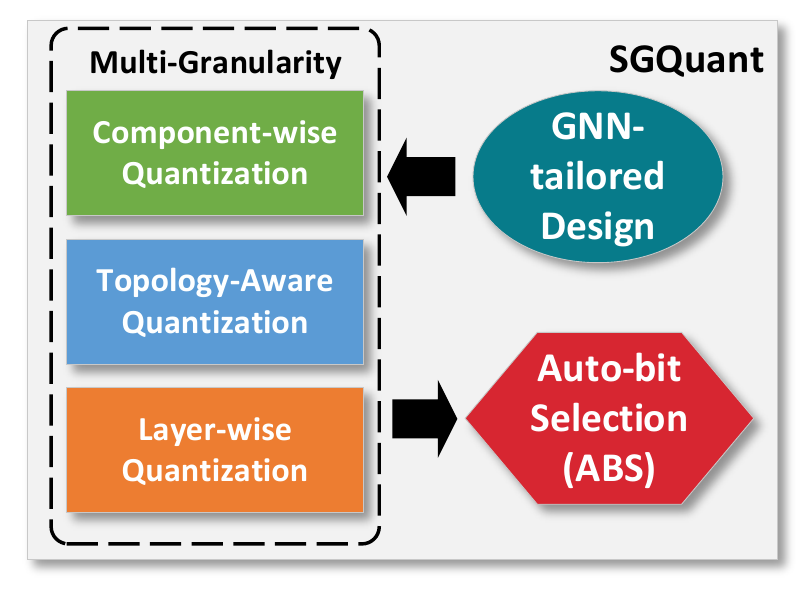}
    \caption{Overview of SGQuant.}
    \vspace{-15pt}
    \label{fig: Overall Architecture.}
\end{figure} 


Based on these observations, we make the following contributions in this paper to systematically quantize GNNs, as illustrated in Figure \ref{fig: Overall Architecture.}.
\begin{itemize}
    \item We propose a GNN-tailored quantization algorithm design to reduce memory consumption and a GNN quantization finetuning to maintain the accuracy.
    \item We propose a multi-granularity quantization featured with \textit{component-wise}, \textit{topology-aware}, and \textit{layer-wise} quantization to meet the diverse data precision demands.
    \item We propose end-to-end bits selecting in an automatic manner that makes the most appropriate choice for the aforementioned different quantization granularities.
    \item Rigorous experiments show SGQuant can reduce the memory up to 31.9$\times$ (from 4.25$\times$) compared with the original full-precision model meanwhile limiting the accuracy to 0.4\% on average.
\end{itemize}

\section{Backgrounds and Related Work}
In this section, we will first introduce the basics of Graph Neural Networks (GNNs), and then give some background knowledge of applying data quantization on GNNs.

\subsection{Graph Neural Network}

Graph Neural Networks (GNNs) are now becoming the major way of gaining insights from the graph structures.
It generally includes several graph convolutional layers, each of which consists of two components: an \textit{Attention Component} and a \textit{Combination Component}, as illustrated in Figure~\ref{fig: node renumbering and block based memory}.
Formally, given two neighboring nodes $u$ and $v$ (\textit{i.e.}, $u \in \mathcal{N}(v)$), and their node embedding $h_u^k \in \mathcal{R}^D$ and $h_v^k \in \mathcal{R}^D$ at layer $k$, GNNs first use the attention component to measure the relationship between these two nodes:
\begin{equation} \label{eq:attention} \small
\alpha_{u,v}^{k}  = Attention({h_{u}^k, h_{v}^k, \mathcal{W}_{att}^k|u\in \mathcal{N}(v)})
\end{equation}
One instantiation of the attention component is a single-layer neural network that concatenates the node embedding $h_{u}^k$ and $h_v^k$, and multiplies with the attention weight matrix $\mathcal{W}_{att}^{k}$, as the case in GAT~\cite{GATConv}. Note that GCN~\cite{GCNConv} is a special case that has the attention weight matrix with all elements equaling to one. Overall, $\alpha^k \in \mathcal{R}^{N\times N}$ is an \textit{attention matrix} measuring the pairwise relationship between nodes, whose memory consumption increases quadratically as the number of nodes $N$ increases.

Then, GNN computes the node embedding $h_v^{k+1}$ for node $v$ at layer $k+1$ with the combination component:
\begin{equation} \small 
    h_{v}^{k+1} = Combination(h_v^k, h_u^k, a_{u,v}^{k}, \mathcal{W}_{com}^{k+1} | u \in \mathcal{N}(v))
\end{equation}
One popular instantiation of the combination component is to 1) average over embeddings from neighboring nodes weighted by the attention matrix $alpha$; and 2) multiply the averaged embedding with a combination weight $W_{com}^{k+1}$:
\begin{equation} \small \label{eq:combination}
     h_v^{k+1} = W_{com}^{k+1} \cdot \sum_{u\in\mathcal{N}(v)} \alpha_{u,v}^k h_u^k
\end{equation}
For each layer $k$, GNNs have $N_k$-dimension embedding vector for each node and an \textit{embedding matrix} $h^{k} \in \mathcal{R}^{N\times N_k}$ when storing all embeddings for $N$ nodes.
This embedding matrix will increase linearly with the number of nodes $N$ and introduce heavy memory overhead for large graphs (\textit{e.g.}, Reddit \cite{snapnets} with $232,965$ nodes).
\begin{figure}[t] \small
    \centering
    \includegraphics[width=\columnwidth]{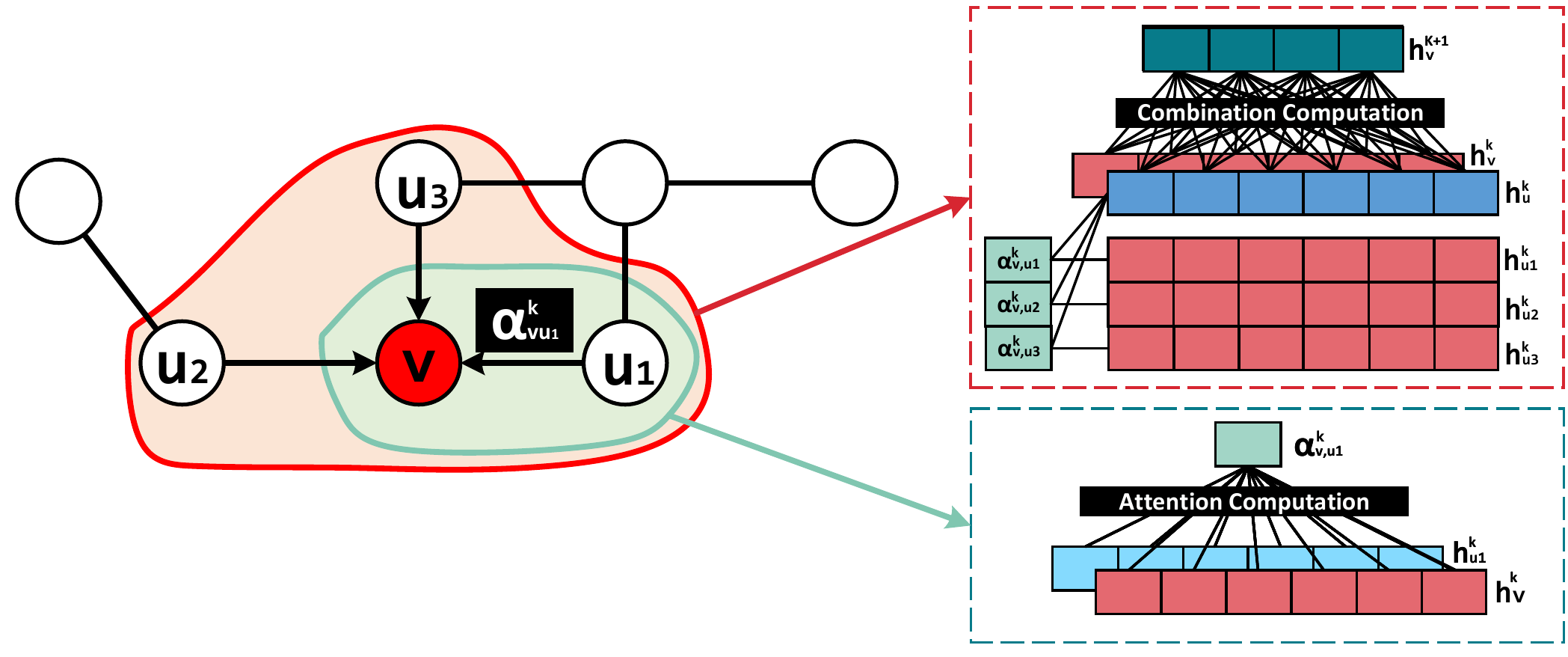}  
    \caption{Example illustrating the standard GNNs. $v$, $u_1$, $u_2$, and $u_3$ stands for four graph vertices and $h_{v}^k$ refers to the feature vector for node $v$ at the $k^{th}$ layer.}
\vspace{-15pt}
    \label{fig: node renumbering and block based memory}
\end{figure}

Besides the concepts of \textit{Layer} and \textit{Component}, GNNs also consider the third concept -- \textit{Topology}. The topology of GNNs characterizes the graph structure based on the properties of nodes and the edge connections among them.

\subsection{Quantization}
Numerous works of quantization mostly focus on data compression around Convolutional Neural Networks (CNNs). Song et al.~\cite{han2015deep} reduces the size of CNNs without accuracy loss by network pruning, weight quantization, post-quantization network fine-tuning.
Ron et al.~\cite{banner2019post} offers a post-training quantization targeting at weights and activations and minimizes memory consumption at the tensor level.
Z. Chen et al.~\cite{zhu2016trained} proposes ternary-value based weight quantization to reduce the size of neural networks with minimal accuracy loss.
Darryl~\cite{lin2016fixed} introduces a layerwise quantizer for fixed-point implementation of DNNs.

Despite such great success in CNNs, efficient GNN quantization is yet to come.
And we believe that GNNs have great potential for quantization. There are several reasons, 1) GNN architectures display different levels of computation hierarchy, which allow for more specialized quantization based on their properties (\textit{e.g.}, different quantization strategies for nodes with different degrees, or for layers with different hidden dimensions), which facilitates a fine-grained quantization scheme based on the types of operations, whereas CNNs have fixed shape of all feature maps that pass through the same set of NN layers; 2) GNNs are more diverse in their inputs, such as graph topologies (\textit{e.g.}, edge connections) and node features (embeddings), which enable quantization based on the categories of the data.
Overall, we are the first to systematically and comprehensively explore the quantization on GNNs by exploiting graph properties and GNN architecture.

\section{GNN-tailored Quantization} \label{sec:uniform}



In this section, we introduce our GNN-tailored quantization to convert a full-precision GNN to a quantized GNN with reduced memory consumption.

\subsection{Quantization Algorithm Design}
Two key designs differentiate our GNN quantization from existing work on CNN quantizations.
First, SGQuant quantizes both the attention matrix $\alpha^k \in \mathcal{R}^{N\times N}$ and the embedding matrix $h^k \in \mathcal{R}^{N\times D_k}$, while the CNN quantization generally only considers feature quantization due to the intrinsic model difference between GNNs and CNNs.
Second, when assigning different quantization bits for the attention matrix and the embedding matrix, SGQuant contains a ``rematching'' mechanism that matches their quantization bits and enables the computation in Equation \ref{eq:combination}.

Formally, given a quantization bit $q$ and the 32-bit attention matrix $\alpha^k \in \mathcal{R}^{N\times N}$ computed from Equation \ref{eq:attention}, we quantize it as a q-bit attention matrix
\begin{equation} \small \label{eq:scale_quantize}
    \alpha^{k,(q)} = \floor*{\frac{\alpha^{k} - \alpha_{min}}{scale}}.
\end{equation}
where $\alpha_{min}$ is an empirical lower bound of the attention matrix values, $scale$ is the ratio between the attention matrix range and the q-bit representation range, and $\floor*{\cdot}$ is the floor function.
Specifically, we evaluate the GNN on large graph benchmarks and collect the statistics on its attention matrices, including the minimal value $\alpha_{min}$, the maximum value $\alpha_{max}$.
Then, we can compute a 32-bit $scale$ parameter on the ratio as the feature range $\alpha_{max}-\alpha_{min}$ over the q-bit representation range $2^q$.
While Equation \ref{eq:attention} generates a 32-bit attention matrix $\alpha^k \in \mathcal{R}^{N\times N}$ requiring $32\times N \times N$-bit memory space, our quantized q-bit attention matrix requires only $q\times N \times N$-bit memory space.
In this way, our quantization on the attention matrix reduces the memory consumption to $q/32$ of its full-precision version.
In particular, once we have computed the 32-bit attention value $\alpha_{u,v}^k$, we can immediately quantize it into a $q$-bit value and store it in the memory.
Similarly, given a quantization bit $p$ and the 32-bit embedding matrix $h^k \in \mathcal{R}^{N\times D_k}$ from Equation \ref{eq:combination}, we can generate a $p$-bit embedding matrix $h^{k,(p)}$ that reduces the memory consumption to $p/32$ of its full-precision version.

Suppose we assign different quantization bits $p$ and $q$ to the attention matrix and the embedding matrix, there would be an ``unmatching bit'' problem in Equation \ref{eq:combination} that the attention value $\alpha_{u,v}^{k,(q)}$ and the embedding $h_u^{k,(p)}$ have unmatching bits.
To solve this problem, we propose a ``rematching'' mechanism that recovers the quantized value to $32$-bit value before these values enter the combination component.
Specifically, we compute the recovered 32-bit attention as $\alpha_{u,v}^{k, (q)'} = scale \cdot \alpha_{u,v}^{k,(q)} + \alpha_{min}$.
Similarly, we can compute the recovered 32-bit embedding $h_v^{k,(p)'}$.
Feeding these recovered values into the combination component, we can compute the Equation \ref{eq:combination} as
\begin{equation} \small \label{eq:rematching}
     h_v^{k+1} = \mathcal{W}_{com}^{k+1} \cdot \sum_{u\in\mathcal{N}(v)} \alpha_{u,v}^{k,(q)'} h_u^{k,(p)'}
\end{equation}
Note that the ``rematching'' mechanism introduces negligible memory overhead since, when we compute the node embedding $h_{v}^{k+1}$, we only recover a small set of nodes $u$ that have edge connections with the node $v$.
In addition, the $\mathcal{W}_{com}^{k+1}$ here is a 32-bit value since SGQuant only quantizes the GNN features as discussed in Figure \ref{fig: Feature/Weight Memroy Size Ratio.}.
Similarly, we can compute the attention component at layer $k$ as
\begin{equation}\small
    \alpha_{u,v}^{k}  = Attention({h_{u}^{k,(p)'}, h_{v}^{k,(p)'}, \mathcal{W}_{att}^k|u\in \mathcal{N}(v)})
\end{equation}
where the $\alpha_{u,v}^k$ is a 32-bit value and can be quantized into $q$-bit values with Equation \ref{eq:scale_quantize}.
Due to this similarity, we will only discuss quantizing the combination component in the following sections.

\subsection{GNN Quantization Finetuning} \label{sec:finetuning}
One challenge in GNN quantization is that directly applying the quantization to GNNs during inference usually leads to high accuracy loss up to $10\%$.
This accuracy loss can be largely recovered to less than $0.5\%$ when we finetune the quantized GNNs.
Note that this finetuning procedure only needs to be conducted once for a quantized GNN model.
Overall, SGQuant uses the same loss as the original GNN model (\textit{e.g.}, negative log-likelihood (NLL) for semi-supervised node classification task).
On the backpropagation related to GNN quantization, we derive the gradient as follows
\begin{equation} \small \label{eq:bp}
\begin{split}
    \frac{\partial L}{\partial \alpha_{u,v}^{k,(q)'}} & = \mathcal{W}_{com}^{k+1} \cdot (\frac{\partial L}{\partial h_v^{k+1}} \cdot h_u^{k,(p)'} + \frac{\partial L}{\partial h_u^{k+1}} \cdot h_v^{k,(p)'} ) \\
    \frac{\partial L}{\partial \alpha_{u,v}^{k}} & = \frac{\partial L}{\partial \alpha_{u,v}^{k,(q)'}} \cdot scale \cdot \frac{\partial\alpha_{u,v}^{k,(q)}}{\partial \alpha_{u,v}^k}
\end{split}
\end{equation}
Note that the computation of $\alpha^{k,(q)}$ uses a floor function  $\floor*{\cdot}$, whose gradient is zero almost-everywhere and hinders the backpropagation.
Our SGQuant uses the straight-through estimator that assigns the gradient $\frac{\partial\alpha_{u,v}^{k,(q)}}{\partial \alpha_{u,v}^k}$ to be $1/scale$.
To this end, we can rewrite the gradient $\frac{\partial L}{\partial \alpha_{u,v}^{k}}$ in Equation \ref{eq:bp} as 
\begin{equation} \small
\begin{split}
    \frac{\partial L}{\partial \alpha_{u,v}^{k}} & = \frac{\partial L}{\partial \alpha_{u,v}^{k,(q)'}} \cdot scale \cdot \frac{\partial\alpha_{u,v}^{k,(q)}}{\partial \alpha_{u,v}^k} \\
    &= \frac{\partial L}{\partial \alpha_{u,v}^{k,(q)'}}
\end{split}
\end{equation}
We implement a tailored GNN quantization layer in PyTorch-Geometric~\cite{pyG} that enables both the quantized inference and the backpropagation, such that SGQuant can easily conduct end-to-end finetuning.





\begin{figure*}\small
  \centering
  \includegraphics[width=0.9\linewidth]{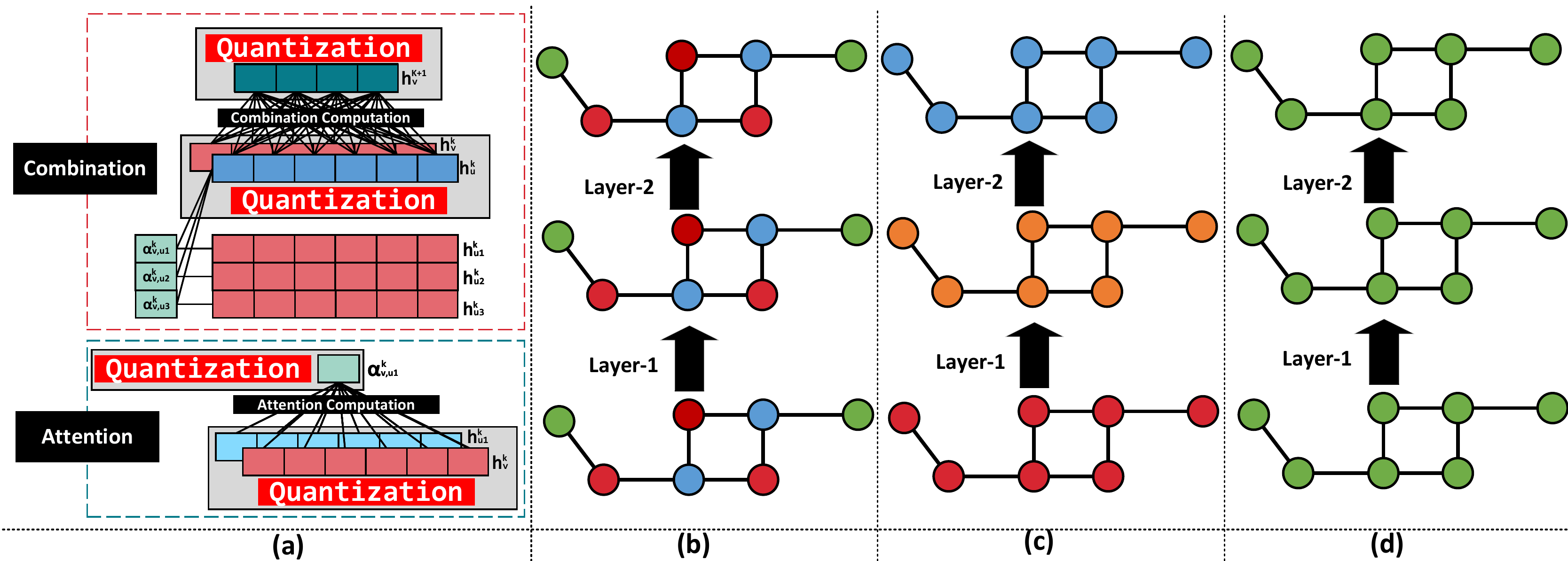}
  \caption{Multi-Granularity Quantization: (a) Component-wise, (b) Topology-aware, (c) Layer-wise, and (d) Uniform Quantization. \small{NOTE: the same color represents the same quantization bit.}
  }
  \vspace{-10pt}
  \label{fig: Multi-Granularity Quantization}
\end{figure*}

\section{Multi-Granularity Quantization}
When designing our specialized graph quantization (SGQuant) method, the quantization granularity is an important aspect to be considered.
In this section, we propose four different types of granularity: \textit{component-wise}, \textit{topology-aware}, \textit{layer-wise}, and \textit{uniform}, as illustrated in Figure~\ref{fig: Multi-Granularity Quantization}.
The simplest granularity is the uniform quantization, which applies the same quantization bits to all layers and components in the GNN.
It helps reduce the memory consumption by replacing the $32$-bit values with the corresponding $q$-bit quantized data representation.
However, when applying the same quantization bit to all layers, nodes, and components, we ignore their different sensitivity to quantization bits and the introduced numerical error, leading to degraded accuracy. To this end, we need the quantization at finer granularity to cater the different sensitivity.

\subsection{Component-wise Quantization} \label{sec:CWQ}
Component-wise Quantization (CWQ) considers the quantization sensitivity at each GNN component and applies different quantization bit to different components, as illustrated in Figure~\ref{fig: Multi-Granularity Quantization}(a).
In each layer, modern GNNs usually contain the attention component for measuring the relationship for each pair of nodes, and the combination component for computing the embedding $h_v^{k+1}$ for the next layer.
While the combination component is critical for providing fine-grained features for the next GNN layer, the attention component usually only provides a coarse-grained hint on the importance of one node $u$ to another node $v$.
Our key insight is that \textit{the attention component is more robust to the numerical error in the GNN quantization compared to the combination component}.
Thus, we can usually apply a lower quantization bit on the attention component than the combination component.

Formally, CWQ maintains a quantization configuration 
\begin{equation} \small
    \{att \; \text{:} \; q_{att}, com \; \text{:} \; q_{com}\}
\end{equation}
for the quantization bits of each GNN component (\textit{i.e.}, $att$ for the aggregation component and $com$ for the combination component), where $q_{att}$ and $q_{com}$ are the quantization bits for the attention and the combination component, respectively.
During the quantization, we will check the quantization bits for each component and conduct quantization correspondingly.
Formally, we compute the quantized attention matrix $\alpha^{k,(q_{att})}$ and the quantized embedding $h^{k,(q_{com})}$, as described in Equation \ref{eq:scale_quantize}.
Note that the $scale$ parameter in Equation \ref{eq:scale_quantize} varies for components according to the assigned quantization bit $q_{att}$ and $q_{com}$.
While CWQ may lead to multiplying two values with "unmatching" bits during the combination component, we can resolve with the "rematching" mechanism in Equation \ref{eq:rematching}.
In particular, during combination, we first recover the quantized component values $\alpha_{u,v}^{k, (q_{att})}$ and $h_u^{k, (q_{com})}$ to their corresponding $32$-bit representation $\alpha_{u,v}^{k, (q_{att})'}$ and $h_u^{k, (q_{com})'}$, then compute with $32$-bit values
\begin{equation} \small
     h_v^{k+1} = W_{com}^{k+1} \cdot \sum_{u\in\mathcal{N}(v)} \alpha_{u,v}^{k, (q_{att})'} h_u^{k, (q_{com})'}
\end{equation}
\subsection{Topology-aware Quantization} \label{sec:topology-aware-quantization}
Topology-aware Quantization (TAQ) exploits the graph topology information and applies different quantization bits for different nodes based on their most essential topology property -- degree, as illustrated in Figure \ref{fig: Multi-Granularity Quantization}(b).
In GNN computation, nodes with higher degrees usually have more abundant information from their neighboring nodes, which makes them more robust to low quantization bits since the random error from quantization can usually be averaged to $0$ with a large number of aggregation operation.
In particular, given a quantization bit $q$, the quantization error $Error_u$ of each node $u$ is a random variable and follows the uniform distribution
$Error_u \sim \mathcal{U}(-\frac{range}{2^q}, \frac{range}{2^q})$
where $range$ represents the difference between the maximum embedding value and the minimum embedding value.
For a node with a large degree, we will aggregate a large number of $Error_u$ and $Error_v$ from the node $u$ and its neighboring nodes $v$ and the average results will converge to $0$ following the \textit{law of large numbers} \cite{shao2003mathematical}.
To this end, nodes with a large degree are more robust to the quantization error and we can use smaller quantization bits to these high-degree nodes.
\begin{figure} \small
  \centering
  \includegraphics[width=\columnwidth]{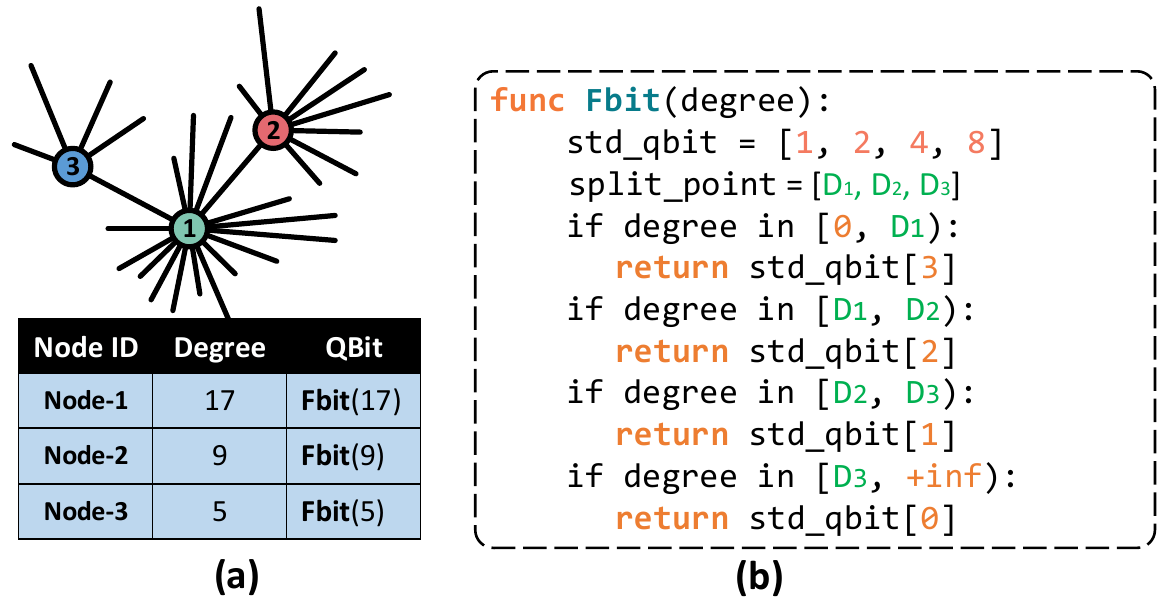}  
  \caption{Topology-aware Quantization.}
  \vspace{-15pt}
  \label{fig:topology}
\end{figure}

Formally, TAQ maintains a quantization configuration that are selected according to the node degrees
\begin{equation} \small
    \{[D_j,D_{j+1}) : q_{Dj} \; | \; j \in [0,1,2,3]\}
\end{equation}
where the features of a node are assigned the quantization bit $q_{Dj}$ if the node degree falls into $[D_j, D_{j+1})$.
Here, we set $D_0=0$ and $D_4 = +\infty$, as illustrated in Figure \ref{fig:topology}.
Suppose there are three nodes: \textit{node-1}, \textit{node-2} and \textit{node-3}, which have the node degree $17$, $9$, and $5$ respectively.
Our TAQ determines the quantization bits of each node based on their degrees.
To get the appropriate quantization bit for different nodes, we propose and implement a $Fbit$ function, as illustrated in Figure~\ref{fig:topology}(b).
We first create the most commonly used quantization bits as a template list ($std_qbit$), and pre-define the degree $split\_point$ list $[D_{1}, D_{2}, D_{3}]$. 
$Fbit$ function maps the nodes with corresponding quantization bits based on their degrees. The strategy behind such a mapping is to maintain higher quantization bits for low-degree nodes, while penalizing high-degree nodes with low bit quantization. 

Once we have assigned different quantization bits $q_u$ to different nodes $u$, there are still ``unmatching'' bits problem across nodes, similar to the ``unmatching'' problem across components.
We can use the ``rematching'' technique on node embeddings and compute the combination component as
\begin{equation} \small
     h_v^{k+1} = \mathcal{W}_{com}^{k+1} \cdot \sum_{u\in\mathcal{N}(v)} \alpha_{u,v}^k h_u^{k, (q_u)'}
\end{equation}
where $\alpha_{u,v}^k$ is a 32-bit value.
TAQ does not quantize the attention matrix since we consider only the first-order topology information and skip second-order topology information that, for an edge $u\-v$, two nodes $u$ and $v$ have different degrees.






\subsection{Layer-wise Quantization} \label{sec:LWQ}
The Layer-wise Quantization (LWQ) exploits the diverse quantization sensitivity in individual GNN layers and provides different quantization bits to each layer.
Our key motivation is that leading layers usually take the detailed data and capture the low-level features while the succeeding layers usually abstract these low-level details into high-level features.
To this end, leading layers require large quantization bits to represent the low-level details while the succeeding layers need only small quantization bits for storing the high-level features.
Our evaluation empirically confirms that, under the same memory consumption, assigning higher bits to the leading layers generally leads to higher accuracy, compared to assigning higher bits to the succeeding layers.



Formally, LWQ maintains a quantization configuration
\begin{equation} \small
    \{k \; \; \text{:} \;\; q_k \; | \; k \in [1,2,...,n]\}    
\end{equation}
where $q_k$ is the quantization bit at the layer $k$ and $n$ is the number of GNN layers.
In particular, GNN quantized with LWQ has the same quantization bits $q_k$ for both the attention matrix $\alpha^k$ and the embedding matrix $h^k$ at the layer $k$ and computes the combination component $h_v^{k+1}$ at $k+1$ as
\begin{equation} \small
     h_v^{k+1} = W_{com}^{k+1} \cdot \sum_{u\in\mathcal{N}(v)} \alpha_{u,v}^{k, (q_k)'} h_u^{k, (q_k)'}
\end{equation}


\subsection{Combine Multiple Granularities}
Besides applying the above granularities stand-alone, SGQuant can effectively combine them in collaborative ways. And we detail two major types of combinations as follows.

\paragraph{\textbf{LWQ+CWQ}}
Note that LWQ and CWQ are complementary and can be easily combined to provide more fine-grained quantization configuration
\begin{equation} \small
    \{(k,att)\text{:}q_{k,att}, (k,com)\text{:}q_{k,com} | k \in [1,2,...,n]\}
\end{equation}
Our evaluation shows that \textit{LWQ+CWQ} can provide lower quantization bits at the same accuracy, compared to only applying LWQ or CWQ alone.
The main insight is that \textit{LWQ+CWQ} provides more fine-grained granularities and could potentially generate models with higher accuracy under the same memory budget.
Formally, GNN quantized with \textit{LWQ+CWQ} computes the combination component as
\begin{equation} \small
     h_v^{k+1} = W_{com}^{k+1} \cdot \sum_{u\in\mathcal{N}(v)} \alpha_{u,v}^{k, (q_{k,att})'} h_u^{k, (q_{k,com})'}
\end{equation}
We can similarly use \textit{LWQ+TAQ} and \textit{CWQ+TAQ}.
We omit the details of these two combinations here due to page limits.



\paragraph{\textbf{LWQ+TAQ+CWQ}}
We can also combine TAQ, LWQ, and CWQ to generate quantization configuration
\begin{equation} \small
\begin{split}
    \{(k,att)\text{:} q_{k,att}, & (k,com, [D_j, D_{j+1}))\text{:} q_{k,com,D_j} \\
    &| k \in [1,2,...,n], j \in [1,..,4]\}
\end{split}
\end{equation}
Note that the quantization bits $q_{k,att}$ on the attention matrix does not depend on the topology information, as the case in TAQ.
Formally, GNN with \textit{LWQ+TAQ+CWQ} computes as
\begin{equation} \small
     h_v^{k+1} = W_{com}^{k+1} \cdot \sum_{u\in\mathcal{N}(v)} \alpha_{u,v}^{k, (q_{k,att})'} h_u^{k, (q_{k,com,D_j})'}
\end{equation}
where $D_j$ is decided by the degree of node $u$.

\section{Auto-bit Selection} \label{sec:selection}
Given the rich set of quantization granularities, one natural question arises:
\textit{How can we assign quantization bits for different granularities to achieve the sweet point between accuracy and memory saving}?
Essentially, we need to solve a combinatorial optimization problem that minimizes the end-to-end loss by selecting a group of discrete quantization bits.
Suppose we are considering \textit{LWQ+TAQ+CWQ}, we can formalize the combinatorial optimization problem as
\begin{equation} \small
\begin{split}
    \min_{\substack{q_{k,att},\\ q_{k,com,D_j}}}
     Loss(\alpha_{u,v}^{k, (q_{k,att})}, h_u^{k,(q_{k,com,D_j})}, W_{com}^{k}, W_{att}^k)
\end{split}
\end{equation}
where $Loss(\cdot, \cdot, \cdot, \cdot)$ is typically cross-entropy for classification tasks and $L_2$ norm for regression tasks, $q_{k,att}$ is the quantization bits for attention matrix at layer $k$, $q_{i,com,D_j}$ is the quantization bits for the node embedding at layer $k$ for node $u$.
We also include the weight matrices $W_{com}^k$ and $W_{att}^k$ in the loss function, since we conduct end-to-end finetuning for the quantized GNN, as discussed in Section \ref{sec:finetuning}.

There are three challenges in solving this combinatorial optimization problem.
First, there is a large design space due to the abundant quantization granularity.
When we apply LWQ, CWQ, and TAQ simultaneously, the number of possible quantization configurations increases exponentially, leading to huge manual efforts in exploration.
Second, large diversity exists in the GNN model design in terms of the attention generation in the aggregation components and the neural network design in the combination components.
This diversity makes it hard to analytically compute the end-to-end quantization error and the impact of quantization bits towards the GNN predictions.
Third, graph topology usually varies in terms of the number of nodes and the degree distribution, making the measurement of quantization intractable.
As we have discussed in Section \ref{sec:topology-aware-quantization}, this topology information usually has a high impact on the quantization error, requiring the consideration of both the graph topology and the GNN design during the selection of quantization bits.

To address these challenges, we build an auto-bit selection (ABS) with two main components: a \textit{machine learning cost model} that predicts the accuracy of the quantized GNN under a given quantization configuration, and an \textit{exploration scheme} to select the promising configurations.

\subsection{Machine Learning Cost Model}
Before we dive into our machine learning (ML) cost model, we will first discuss two baseline approaches.
The first one is the random search with trial-and-error that randomly samples a large number of quantization configurations and examines all samples to find the best one.
However, this approach usually requires a large number of samples to find a good configuration.
The second is to build a pre-defined cost model to analyze the impact of quantization bits over the predictions for a particular GNN model and a graph topology.
However, this approach usually fails to generalize well to various GNN models and graph inputs.
\begin{figure}[t] \small
    \centering
    \vspace{-8pt}
    \includegraphics[width=\columnwidth]{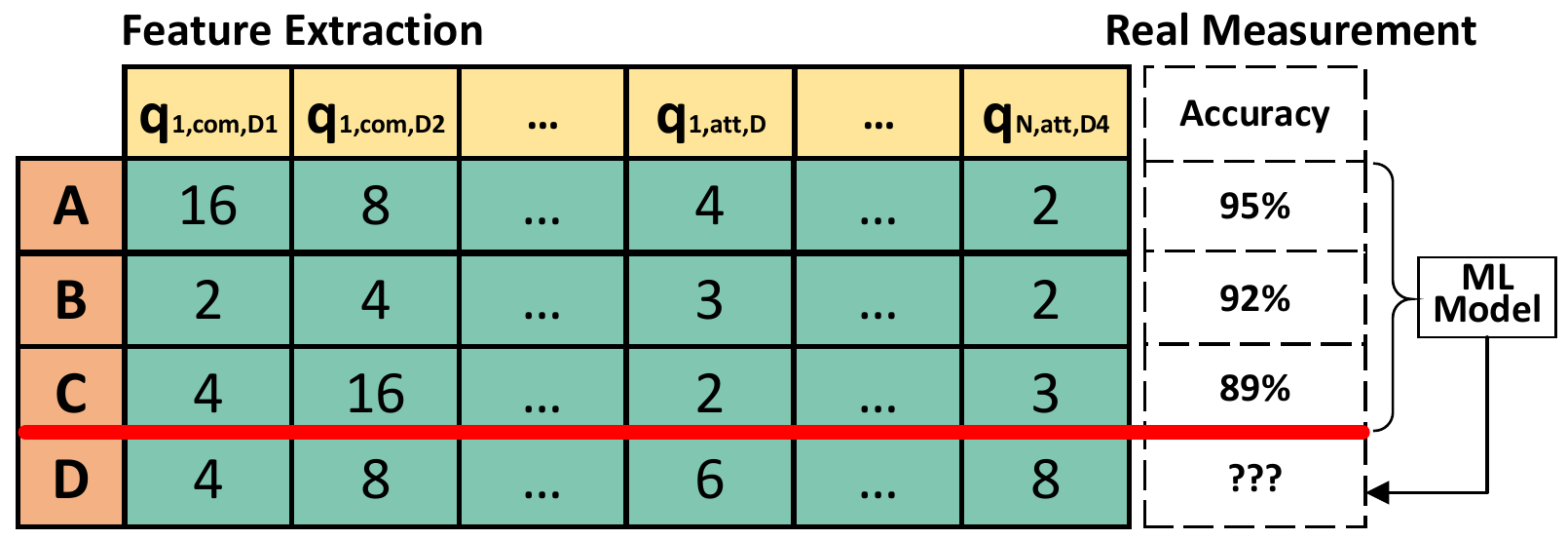}
    \vspace{-5pt}
    \caption{Machine Learning Cost Model.} 
    \vspace{-15pt}
    \label{fig: Machine Learning Cost Model}
\end{figure} 

To this end, we build a ML cost model that learns on the fly the interaction among \textit{quantization bits}, \textit{GNN models}, and the \textit{graph topology}.
Figure~\ref{fig: Machine Learning Cost Model} illustrates our ML cost model design.
Given the quantization granularity and the bits to select, we can randomly generate a set of configurations and extract the quantization bits as the features. 
Then, we will train and evaluate these configurations, and measure the accuracy as true labels.
Finally, we will use the collected features and labels to train our ML cost model and use it to predict the remaining configurations.
We treat this task as a regression problem and use a traditional ML model --- regression tree \cite{Loh2011ClassificationAR}, as our ML cost model.
We prefer the regression tree over neural networks since the former one has faster inference speed and does not require a large amount of training data.

\subsection{Exploration Scheme}

Given the ML cost model, a simple exploration scheme would evaluate all remaining quantization configurations and select the one with the highest predicted accuracy and the lowest memory size.
However, this approach may fail in two cases.
First, it is time-consuming to evaluate all remaining quantization configurations, especially when we use LWQ, CWQ, and TAQ simultaneously for a large GNN.
Second, we may select a small number of quantization configurations for training the ML cost models to reduce the overhead from auto-bit selection, such that the trained ML model cannot predict precise accuracies for all remaining quantization configurations.




To this end, we propose an exploration scheme that iteratively trains the ML cost model and selects promising configurations.
In this way, we can balance the low overhead in training the ML cost model and the precise prediction of configuration accuracies.
In particular, there are five steps in our exploration scheme.
\begin{itemize}
    \item \textbf{Step1}: Randomly select a small number $N_{mea}$ of configurations, extract features, and measure their accuracies.
    \item \textbf{Step2}: Train the ML cost model based on the collected features and labels.
    \item \textbf{Step3}: Sample a large number $N_{sample}$ of configurations, use the ML cost model to predict their accuracy, and find the ones with the $\text{top-}N_{mea}$ accuracy.
    \item \textbf{Step4}: Extract features of the selected configurations and measure their accuracies. 
    \item \textbf{Step5}: Repeat \textbf{Step2} - \textbf{Step4} until reaching $N_{iter}$ iterations.
\end{itemize}

During this procedure, only configurations with negligible accuracy drop ($<0.5\%$) will be kept.
Among the remaining configurations, we select the one with the lowest memory consumption.
Here, $N_{mea}$, $N_{iter}$, and $N_{sample}$ are hyper-parameters in ABS that balance the selection overhead and the ML cost model accuracy.
Smaller $N_{meausre}$ and $N_{iter}$ lead to lower selection overhead by reducing the number of quantization configurations that are trained and evaluated.
We have experimented with diverse $N_{mea}$ and $N_{iter}$ on an extensive collection of GNNs and datasets, and find that a small $N_{measure}=40$ and a small $N_{iter}=5$ hit the balance between selection overhead and the ML cost model accuracy.
The reason is that our cost model is a traditional regression tree model that can be trained with a small amount of data.
Using a larger $N_{sample}$, we can generally select configurations with lower memory consumption and higher accuracy, since more configurations are evaluated by our ML cost model.
By default, we set $N_{sample}=2000$ in our evaluation.
This leads to negligible latency ($<0.1$ seconds) at each iteration due to the fast inference speed from the regression tree.


\begin{table}[t] \small
    \centering
    \caption{GNN Architectures.}
    \label{tab:models}
    \scalebox{1.1}{
    \begin{tabular}{c|c}
    \Xhline{2\arrayrulewidth}
        \textbf{Arch} & \textbf{Specification} \\
    \Xhline{2\arrayrulewidth}
         GCN & hidden=32, \#layers=2 \\
    \hline
         AGNN & hidden=16, \#layers=4 \\
    \hline
        GAT & hidden=256, \#layers=2 \\
    \Xhline{2\arrayrulewidth}
    \end{tabular}
    }
\end{table}

\begin{table}[t] \small
    \caption{Datasets for Evaluation.}
    \label{tab:dataset}
    \centering
    \scalebox{1}{
     \begin{tabular}{ l c c c c }
    \Xhline{2\arrayrulewidth}
    \textbf{Dataset} & \textbf{\#Vertex} & \textbf{\#Edge} & \textbf{\#Dim} & \textbf{{\#Class}}\\
    \Xhline{2\arrayrulewidth}
    Citeseer    & 3,327	    & 9,464	    & 3,703 & 6      \\
    Cora	    & 2,708     & 10,858	& 1,433 & 7      \\
    Pubmed	    & 19,717	& 88,676	& 500  & 3      \\
    \hline
    Amazon-computer & 13,381    & 245,778	    & 767  & 10 \\
    Reddit        & 232,965     & 114,615,892   &  602 & 41\\
    \Xhline{2\arrayrulewidth}
    \end{tabular}
    }
    \vspace{-10pt}
\end{table}

\begin{table*}[t] \small
    \centering
    \caption{Overall Quantization Performance.}
    \scalebox{0.9}{
    \begin{tabular}{|c|l|c|cc|c|}
    \hline
        \textbf{Dataset} & \textbf{Network} & \textbf{Accuracy (\%)}  &\textbf{Average Bits} &\textbf{Memory Size (MB)} & \textbf{Saving} \\
    \hline
    \hline
        \multirow{6}{*}{Cora}&GCN (Full-Precision) & $82.2$& $32$ & $15.42$ & - \\
        &GCN (Reduced-Precision)& $81.72$ & $1.22$ & $0.59$ & $26.1\times$\\
        &AGNN (Full-Precision)& $83.16$ & $32$ & $15.94$ & - \\
        &AGNN (Reduced-Precision)& $82.75$ & $2.15$ & $1.07$ & $14.90\times$\\
        &GAT (Full-Precision)& $82.50$ & $32$ & $16.21$ & -\\
        &GAT (Reduced-Precision)& $82.10$ & $2.58$ & $1.31$ & $12.37\times$\\
    \hline
    \hline
        \multirow{6}{*}{Citeseer}&GCN (Full-Precision) & $71.82$ & $32$ & $51.06$ & -\\
        &GCN (Reduced-Precision)& $71.54$ & $1.01$ & $1.6$ & $31.9\times$ \\
        &AGNN (Full-Precision)& $71.58$ & $32$ & $50.01$ &-\\
        &AGNN (Reduced-Precision)& $71.18$ & $1.08$ & $1.69$ & $29.59\times$\\
        &GAT (Full-Precision)& $71.10$ & $32$ & $59.49$ & - \\
        &GAT (Reduced-Precision)& $70.70$ & $2.42$ & $3.82$ & $13.2\times$\\
    \hline
    \hline
        \multirow{6}{*}{Pubmed}&GCN (Full-Precision) & 80.36 & 32 & 43.71 & -\\
        &GCN (Reduced-Precision)& 80.28 & 2.9 & 4.01 & $10.9\times$\\
        &AGNN (Full-Precision)&80.44&32&43.46&-\\
        &AGNN (Reduced-Precision)& 80.31 & 3.07 & 4.17 & $10.42\times$ \\
        &GAT (Full-Precision)&78.00& 32 & 44.48 & -\\
        &GAT (Reduced-Precision)&77.30& 3.77 & 5.26 & $8.47\times$\\
    \hline
    \hline
        \multirow{6}{*}{Reddit}&GCN (Full-Precision) & 81.07 & 32 & 328.70 & -\\
        &GCN (Reduced-Precision)& 80.36 & 3.72 & 38.25 & $8.59\times$\\
        &AGNN (Full-Precision)   & 74.63 & 32 & 643.92 &-\\
        &AGNN (Reduced-Precision)& 74.40 & 4 & 113.92 & 5.65x\\
        &GAT (Full-Precision)&92.66&32&311.85&-\\
        &GAT (Reduced-Precision)&92.23& 4.07 & 39.70 & $7.86\times$\\
    \hline
    \hline
        \multirow{6}{*}{Amazon-Computer}&GCN (Full-Precision) & 89.57 & 32 & 44.58 & -\\
        &GCN (Reduced-Precision)& 89.39 & 3.29 & 4.59 & $9.72\times$\\
        &AGNN (Full-Precision)& 77.69  & 32& 44.16 &-\\
        &AGNN (Reduced-Precision)& 77.33 & 4& 5.99 & $7.37\times$\\
        &GAT (Full-Precision)& 93.10 &32& 45.71 & -\\
        &GAT (Reduced-Precision)& 92.60 & 7.53 & 10.75 & $4.25\times$ \\
    \hline
    \end{tabular}
    }
    \vspace{-5pt}
    \label{tab: Overall Quantization Performance (Accuracy, Average Bits, and Memory Saving)}
\end{table*}

\section{Evaluation} \label{sect: Evaluation}


In this section, we show the strength of our proposed quantization method through intensive experiments over various GNN models and datasets.

\subsection{Experiment Setup}
\subsubsection{\textbf{GNN Architectures}}
\underline{\textbf{Graph Convolutional Network}} (\textbf{GCN})~\cite{GCNConv} is the most basic and popular GNN architecture. It has been widely adopted in node classification, graph classification, and link prediction tasks. Besides, it is also the key backbone network for many other GNNs, such as GraphSage~\cite{SageConv}, and Diffpool~\cite{diffpool}.
\underline{\textbf{Attention-based Graph Neural Network}} (\textbf{AGNN})~\cite{GINConv} aims to reduce the parameter size and computation by replacing the fully connected layer with specialized propagation layers.
\underline{\textbf{Graph Attention Network}} (\textbf{GAT})~\cite{GATConv} is a reference architecture for many other advanced GNNs with more edge properties, which can provide state-of-the-art accuracy performance on many GNN tasks. Details of their configurations are shown in Table~\ref{tab:models}.

\subsubsection{\textbf{Datasets}}
We select two categories of graph datasets to cover the vast majority of the GNN inputs. 
The first type includes the most typical datasets (\textit{Citeseer}, \textit{Cora}, and \textit{Pubmed}) used by many GNN papers~\cite{GCNConv, GINConv, SageConv}. 
They are usually small in the number of nodes and edges, but come with high-dimensional feature embedding.
The second type (\textit{Amazon-computer}, and \textit{Reddit}) are large graphs~\cite{snapnets, GCNConv} in the number of nodes and edges.
Details of the above datasets are listed in Table~\ref{tab:dataset}.

\subsection{Overall Performance}
In this section, we demonstrate the benefits of SGQuant by evaluating its impact of accuracy loss and memory saving.  As shown in Table~\ref{tab: Overall Quantization Performance (Accuracy, Average Bits, and Memory Saving)}, our specialized quantization method can effectively reduce the memory consumption up to $31.9\times$, $29.59\times$, and $13.2\times$ on GCN, AGNN, and GAT respectively, meanwhile limiting the accuracy loss by $0.34\%$,  $0.31\%$, $0.47\%$ on average compared with the original full-precision model for GCN, AGNN, and GAT.

Moreover, there are several noteworthy observations. 
\underline{Across different datasets}: On datasets with smaller sizes, such as \textit{Cora} and \textit{Citeseer}, our specialized quantization method can reduce the memory size more aggressively while maintaining accuracy by selecting relatively low average bits, such as 1.22 for GCN on Cora.
This is because the smaller datasets with limited size of nodes and edge connections make the quantization precision loss less significant.
\underline{Across different models}: we find that to maintain the accuracy, SGQuant would select higher average bits for more complex models. 
For example, on \textit{Amazon-computer} dataset, GAT model locates 7.53 as the average bit, while the AGNN and GCN locate 4 bit and 3.29 bit, respectively.
We observe similar pattern on all other datasets that we evaluated.
The major reason is that more complex GNN models would involve more intricate computations that would easily enlarge the accuracy loss of quantization and require higher bits to offset such loss.
For instance, GAT has to first compute neighbor-specific attention values and scale them with the number of attention heads before the combination component.
Instead, AGNN and GCN have simpler combination component that requires much less effort in computation, getting its loss of quantization well under-controlled even with lower bits.

What also worth mentioning is that on datasets with large size, such as \textit{Reddit}, the absolute memory size saving is significant, which reduces up to $530$MB memory occupation. This can also demonstrate the potential of SGQuant to make the GNNs happen on memory-constrained device more easily.

\subsection{Breakdown Analysis of Multi-granularity Quantization}
In this experiment, we break down the benefits of multi-granularity quantization. Specifically, we apply GAT on Cora. We first evaluate the performance of uniform quantization (Uniform) and layer-wise quantization (LWQ).
Then, we evaluate more fine-grained granularity by combining LWQ with component-wise quantization (CWQ) and apply different quantization bits to individual components at each layer.
For example, for GAT with 2 layers and 2 components of aggregation and combination at each layer, the quantization configuration with LWQ+CWQ has $4$ quantization bits (\textit{i.e.}, $q_{1,agg}, q_{1,com}, q_{2,agg}, q_{2,com}$).
We additionally impose the topology-aware quantization (TAQ) to study the performance of SGQuant when considering LWQ, CWQ, and TAQ, simultaneously.

Figure~\ref{fig: Breakdown Analysis of Multi-granularity Quantization.} shows the error rate of each quantization granularity at each memory size.
Specifically, Uniform shows the highest error rate under each memory size.
This error rate increases significantly when we compress the model to be smaller a certain size (2.5MB).
Compared with Uniform, LWQ achieves lower error rate, due to the flexibility in selecting different bits for different layers.
Moreover, we observe that \textit{LWQ+CWQ} further mitigates such accuracy degradation when reducing the model memory footprint aggressively.
The reason is that \textit{LWQ+CWQ} takes the properties of different layers and different components in to consideration, which can strike a good balance between the memory saving and the accuracy. 
Finally, this experiment also shows that, by incorporating the node information (degree) with \textit{LWQ+CWQ+TAQ}, our SGQuant can achieve even lower error rate at each memory size.
The major reason is that high-degree nodes would intrinsically gather more information from its neighbors compared with the nodes with limited number of neighbors.
In other words, applying more aggressive quantization on high-degree nodes would cause minor information loss.

\begin{figure}[t] \small
    \centering
    \vspace{-5pt}
    \includegraphics[width=0.9\linewidth]{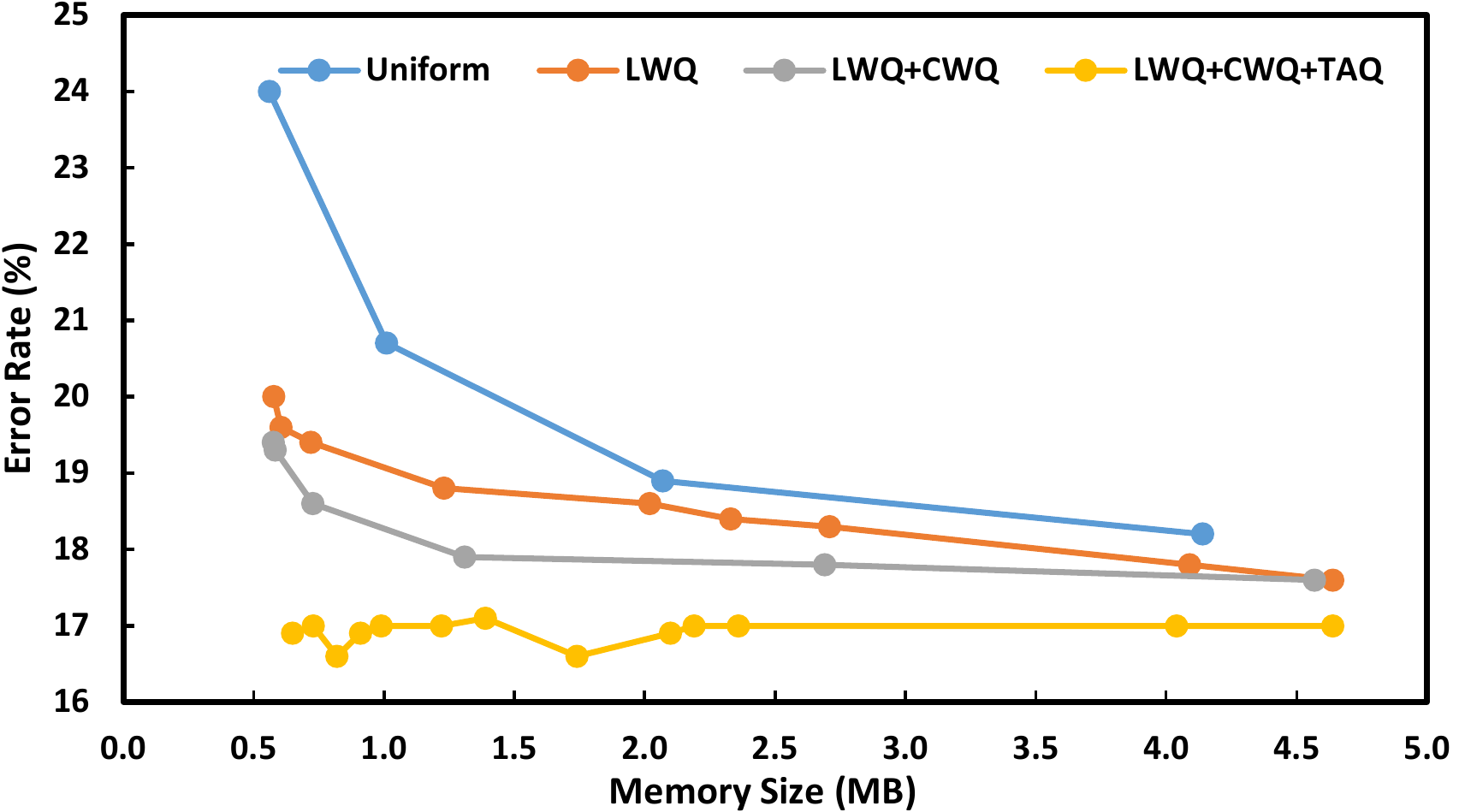}
    \caption{Breakdown Analysis of Multi-granularity Quantization.}
    \vspace{-15pt}
    \label{fig: Breakdown Analysis of Multi-granularity Quantization.}
\end{figure}

\newcolumntype{C}[1]{>{\centering\arraybackslash}p{#1}}
\begin{table}[h] \small
    \centering
    \caption{Optimal Quantization Bit of GAT on Cora.}
    \begin{tabular}{|| >{\centering\arraybackslash}m{6em} | >{\centering}m{11em} | >{\centering\arraybackslash}m{3em}||}
        \hline
        \textbf{Quantization Method} & \textbf{Configuation@Mem-Size=2MB} & \textbf{Error Rate}    \\
        \hline
        \hline
        Uniform &  $q=4$ & 18.90\% \\
        \hline
        LWQ & $q_1=4, \;\; q_2=1 $  & 18.60\% \\
        \hline
        LWQ + CWQ & $q_{1,att} = 2, \;\; q_{1,com}=4$ $q_{2,att}=2, \;\; q_{2,com}=2$   & 17.90\% \\
        \hline
        LWQ + CWQ + TAQ & $q_{1,Dj} = [4,3,2,1]$ $q_{2,Dj} = [1,1,1,1]$ & 16.70\% \\
        \hline        
    \end{tabular}
    \label{tab:optimal-bitwidth1}
\end{table}

As a case study, Table \ref{tab:optimal-bitwidth1} shows the allocated bit-width and error rate of GAT on Cora with different granularity, with the memory size around 2MB.
We observe similar trend as Figure \ref{fig: Breakdown Analysis of Multi-granularity Quantization.} that fine-grained granularities generally lead to lower error rate at a given memory size.
One interesting observation is that LWQ achieves a lower error rate than Uniform, while LWQ chooses lower quantization bit than Uniform at layer 2.
The insight is that the low quantization bit may introduce the regularization effect and prevent overfitting in the training procedure.
Also, LWQ usually assigns higher bits to leading layers, as discussed in Section \ref{sec:LWQ}.
For the \textit{LWQ+CWQ}, we assign smaller quantization bits to the attention component, since attention component is more robust to the numerical error in the GNN quantization, as discussed in Section \ref{sec:CWQ}.
The most fine-grained granularity is \textit{LWQ+CWQ+TAQ}, where we can reduce error rate by $2.2\%$ under the same memory size, compared with the uniform quantization.

\subsection{Effectiveness of Auto-Bit Selection}
In this experiment, we evaluate auto-bit selection (ABS) with the machine learning (ML) cost model.
As discussed in Section~\ref{sec:selection}, we iteratively select and evaluate quantization configurations.
Among these evaluated quantization configurations, we only select configurations that shows negligible accuracy drop ($<0.5\%$) compared to the full-precision GNN.
Among remaining models, we exhibit its memory saving compared to the full-precision GNN.
We compare our ABS with the random search approach, which randomly picks 200 quantization configurations and selects the one with lowest memory size while also showing negligible accuracy drop. 
\begin{figure}[t] \small
    \centering
    \vspace{-5pt}
    \includegraphics[width=0.9\linewidth]{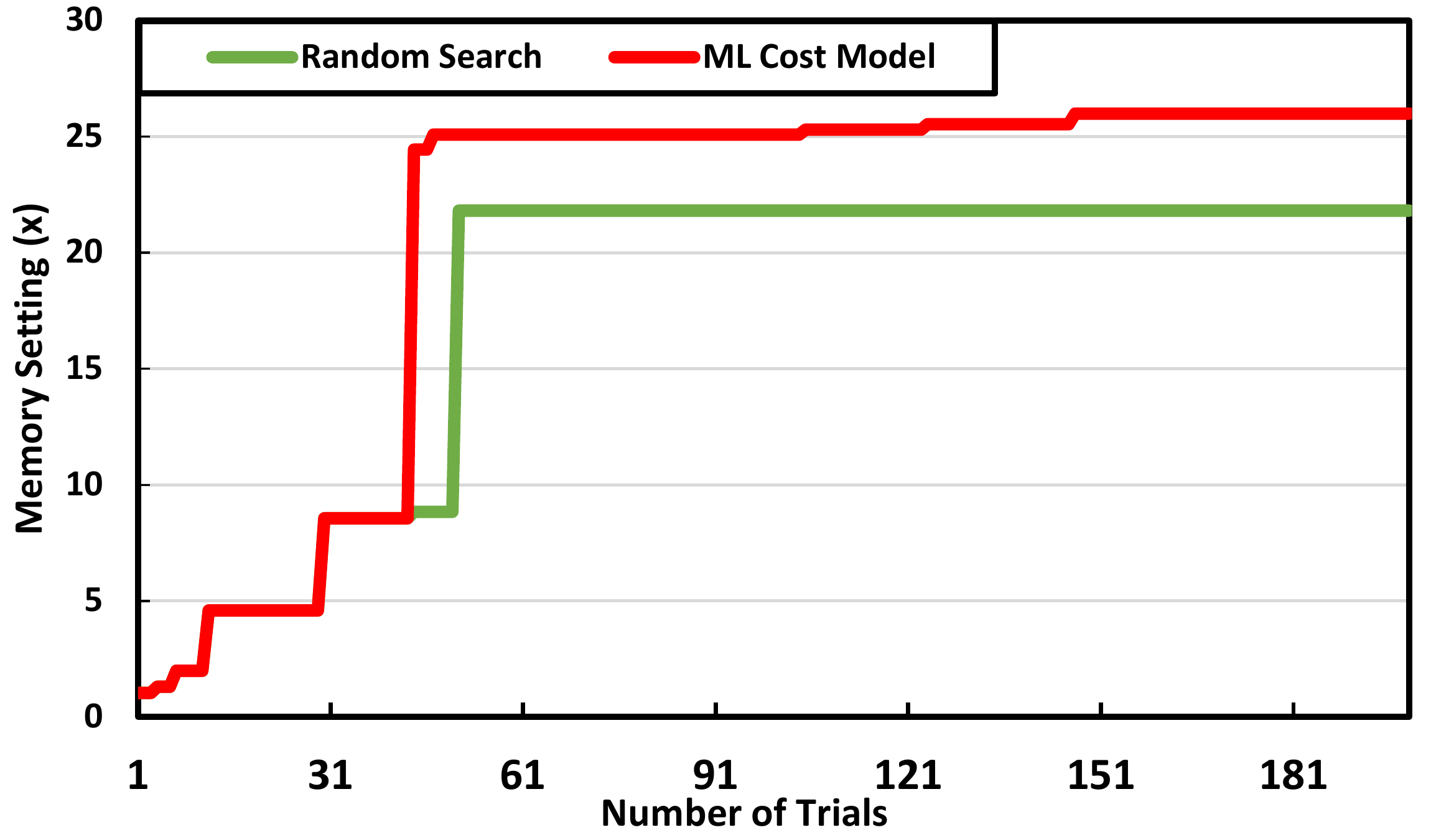}
    \caption{Benefit of ML Cost Model.}
    \vspace{-20pt}
    \label{fig: ml-cost-model-result}
\end{figure}

Figure \ref{fig: ml-cost-model-result} exhibits the results on AGNN and Cora dataset, while similar trend can be observed on other GNNs and datasets.
Overall, our ML cost model converges within 200 trails of quantization configurations and achieves two advantages over the random search approach.
First, ML cost model can locate the appropriate quantization bits more swiftly compared with naive random search solution.
Second, for the final results, ML cost model can pinpoint a more "optimal" value for bits that offers higher memory saving (25$\times$) compared with random search ($20\times$).
The major reasons behind such a success are two folds.
First, we build our initial model based on several key features (configuration parameters) of SGQuant, which can effectively capture the core relation between their value and the final quantization performance (accuracy).
Second, the ML cost model are iteratively updated as it sees more data samples, which helps it refine itself by providing solution more wisely. 
Besides, we observe similar performance between ML cost model and random search at the first $40$ trails.
The reason is that, starting with no training data, our ABS randomly samples and profiles $N_{sample}$(=40) configurations at the beginning, where we can expect similar performance as the random search approach.



\vspace{5pt}
\section{Conclusion}
In this paper, we propose and implement a specialized GNN quantization scheme, SGQuant, to resolve the memory overhead of GNN computing. Specifically, our multi-granularity quantization incorporates the layer-wise, component-wise, and topology-aware quantization granularities that can intelligently compress the GNN features while minimizing the accuracy drop.
To efficiently select the most appropriate bits for all these quantization granularities, we further offer a ML-based automatic bit-selecting (ABS) strategy that can minimize the users' efforts in design exploration.
Rigorous experiments show that SGQuant can effectively reduce the memory size up to $31.9\times$ under negligible accuracy drop.
In sum, SGQuant paves a promising way for GNN quantization that can facilitate their deployment on resource-constraint devices.
}





\bibliographystyle{IEEEtran}
\bibliography{reference}
\end{document}